\documentclass[11pt,twocolumn,table]{article} 
\usepackage{ACL2023}
\usepackage{times}
\usepackage{latexsym}
\usepackage[T1]{fontenc}
\usepackage[utf8]{inputenc}
\usepackage{microtype}
\usepackage{inconsolata}
\usepackage{graphicx}
\usepackage{hyperref}
\usepackage{booktabs}
\usepackage{siunitx}
\usepackage{caption}
\usepackage{arydshln}
\usepackage{pifont}
\usepackage{multirow,multicol}
\usepackage{tcolorbox}
\tcbuselibrary{}

\definecolor{b1}{RGB}{208, 225, 243}
\definecolor{b2}{RGB}{111, 168, 219}
\definecolor{g1}{RGB}{147, 196, 123}
\definecolor{r1}{RGB}{220, 102, 107}

\newcommand{\cmark}{\textcolor{g1}{\ding{51}}} %
\newcommand{\xmark}{\textcolor{r1}{\ding{55}}}   %

\newcommand{\name}{\textsc{DiaLong}}

\title{\name: Benchmarking Reminiscence in Long-Context Dialogues}

\author{Owen Bianchi\thanks{\ \ Equal contribution.} \quad Samer Aslan\footnotemark[1] \quad Ryunosuke Saito\footnotemark[1]\\
Center for Language and Speech Processing, Johns Hopkins University
}

\begin{document}
\maketitle
\begin{abstract}

How well can transformer-based Large Language Models (LLMs) retain and recall memories in long contexts of dialogue?  LLMs are known to forget long‑range factual information in their context, yet the field lacks a comprehensive benchmark to quantitatively assess these shortcomings.  We present \name, a long-context dialogue dataset and accompanying memory task that standardizes evaluation in this space.  It consists of 100 unique conversations created by humans and then augmented with LLMs, extending beyond lengths found in existing datasets.  Our selected memory task is fact recall: classify 40 candidate facts about a dialogue as true or false.  This task directly probes the model's ability to actively retain and retrieve information, serving as a critical measure of memory within extended conversational contexts.  In testing both the latest LLMs and proposed memory enhancements on \name, we demonstrate that more work needs to be done to enable LLMs with better memory performance. By enhancing memory capabilities, LLMs can improve in maintaining coherence, relevance, and human-like interaction over extended conversation. Improving these capabilities will help LLMs fulfill diverse user needs and expectations in various conversational scenarios.

\end{abstract}

\begin{figure}[ht]
\centering
\includegraphics[width=\linewidth]{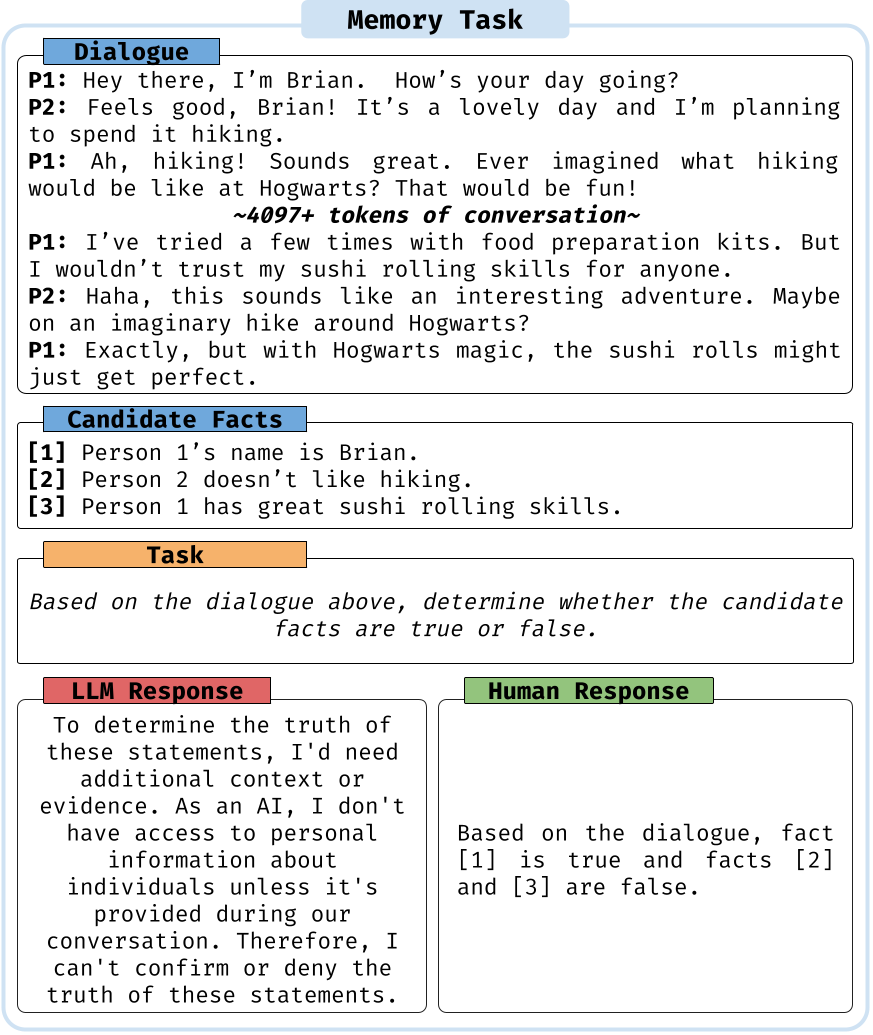}
\caption{An example memory task depicting the performance differences between an LLM and a human in long contexts.}
\label{fig:fig1}
\end{figure}

\section{Introduction}

Large Language Models (LLMs) are constrained by limited context length, which hinders their ability to perform natural language processing tasks. They are deployed in various contexts and applications \cite{brown2020language}, one notable example being conversational agents, which have recently seen a surge in popularity \cite{openai2022chatgpt}\cite{openai2023gpt4}\cite{touvron2023llama}.  Chatbots powered by LLMs are conversational tools capable of understanding and responding to human language effectively.  The unique application of LLMs in chatbots, as opposed to other NLP tasks, lies in their inherent requirement to mimic and produce human-like interactions during dialogues \cite{BrandtzaegFolstad2017Chatbots} \cite{JohariNohuddin2021ChatbotQuality}.  Memory is an essential part of human conversation, allowing for context retention, coherent responses, and building relationships over time.  As defined by Zlotnik and Vansintjan, memory is the process of acquiring, storing, retaining, and retrieving information \cite{zlotnik2019memory}.  LLM-powered chatbots inherently lack long-term memory when compared to humans, hindering their ability to maintain human-like conversation.

Forgetfulness, the tendency of an LLM to lose track of earlier information, is a major limitation. This problem is exacerbated with long contexts, as anything bumped outside of the context window is immediately forgotten.  Directly from OpenAI's API documentation, tokens that do not fit within a model's limit are lost \footnote{``If a conversation has too many tokens to fit within a model’s maximum limit (e.g., more than 4096 tokens for gpt-3.5-turbo) you will have to truncate, omit, or otherwise shrink your text until it fits. Beware that if a message is removed from the messages input, the model will lose all knowledge of it.`` \cite{openai2023gpt4}}.  This is one of the most significant challenges faced by modern LLMs, their performance in long contexts.  It is clear that the limited context window is the main culprit of forgetfulness, as depicted in Figure \ref{fig:fig1}.  A popular approach to address this problem is extending the attention window \cite{vaswani2023attention} \cite{chen2023extending} \cite{su2023roformer}.  The premise is that by enabling a model to process more extensive contexts during inference, it enhances its capability to store, retain, and recall information, thereby extending its memory reach.  This approach is limited for two reasons.  Firstly, contexts in general, especially conversations, can carry on for extremely long periods of time, which would require an impractical length context window.  This is infeasible because extremely long context windows demand excessive computational resources \cite{chen2023longlora}, stemming from the quadratic complexity of self-attention computation \cite{vaswani2023attention}.  Additionally, it has been shown that merely extending the context window results in a considerable degradation in performance for longer contexts \cite{liu2023lost}.

Another culprit for the lack of performance with conversational data is that dialogue presents another problem to LLMs.  Conversational problems are uniquely challenging in that the corpus of knowledge from which the LLM must use to reply grows continuously.  It requires more than just memorization of a training dataset to do well on such tasks, it additionally requires a clever use of the never-before-seen dialogue between the chatbot and the user during inference time.  An approach to address both this and the limited context window is to augment the LLMs with a memory solution such as those discussed in Section ~\ref{sec:relatedwork}. These solutions appear to work at the surface, but we demonstrate that they do not evaluate as rigorously as they should.  Current conversational datasets and methods of evaluation do not focus on both long contexts and memory, and are not a good fit for evaluation of such conversational models.

We introduce \name, a long-context dialogue dataset accompanied by a novel memory benchmark, specifically designed to rigorously evaluate the capacity of current LLMs and emerging memory solutions to sustain and retrieve information across prolonged dialogues. This endeavor underscores a crucial gap in existing methodologies, which predominantly assess models using considerably shorter dialogues, insufficient to capture the intricacies of extensive conversational contexts. Our comprehensive evaluation reveals a pressing need for advancing research to enhance LLMs' memory functions, emphasizing that significant improvements are imperative for these models to achieve and sustain coherence, relevance, and human-like interactions over lengthy conversations. Through \name, we aim to catalyze the development of LLMs that are not only more adept at handling long-context dialogues but also capable of delivering outputs that are coherent, accurate, and trustworthy, thereby better meeting the complex needs and expectations of users in diverse conversational scenarios.

\begin{table*}[htbp]
\centering
\small
\renewcommand{\arraystretch}{1.5}
\begin{tabular}{@{}lccccc@{}}
\toprule
& \textbf{PersonaChat} & \textbf{DailyDialog} & \textbf{EmpatheticDialogues} & \textbf{MultiSessionChat} & \textbf{DiaLong} \\ 
& \cite{zhang2018personalizing} & \cite{li2017dailydialog} & \cite{rashkin2019towards} & \cite{xu2021goldfish} & (this work)\\
\midrule
\rowcolor{b1} Human Written & \cmark & \cmark & \cmark & \cmark & \cmark \\
\rowcolor{b2} Human Validated & \xmark & \xmark & \xmark & \xmark & \cmark \\
\rowcolor{b1} Memory Task & \xmark & \xmark & \xmark & \cmark & \cmark \\
\rowcolor{b2} Dialogue Quality & Low & Variable & High & Low & High \\
\hdashline
\rowcolor{b1} Avg Utt Per Dialogue & 14.7 & 3.9 & 2.6 & 53.3 & 110.7 \\
\rowcolor{b2} Avg Utt Length & 11.9 & 14.5 & 15.3 & 23.0 & 34.1 \\
\bottomrule
\end{tabular}
\captionsetup{justification=centering}
\caption{Comparison of \name~ to popular dialogue datasets}
\label{tab:dialogue_datasets}
\end{table*}

\section{Related Work}
\label{sec:relatedwork}

Relevant literature can be separated into a few main categories. We highlight and discuss prior work on the problem of a limited-context window for LLMs. We then discuss some of the work that has been done towards solving these issues. First, we highlight research focusing on the performance of models on a variety of tasks in long contexts. Next, we cover the main themes and proposed solutions which enable increased memory capabilities in long contexts, especially those beyond the context window. We then mention existing dialogue datasets and show that while they are great for their specific scenarios, a new dataset is needed to evaluate memory abilities in long-context settings. This brings us to the creation of \name.

\paragraph{Forgetting in long contexts.}
The motivation behind our work stems from other studies exploring LLMs' tendency to forget in long contexts. The paper `Challenges and Applications of Large Language Models` \cite{kaddour2023challenges} delves into the overarching challenges faced by LLMs, including a dedicated section on limited context windows. They identify that tackling NLP tasks requires a comprehensive understanding of the entire context, not just isolated words or sentences.  Li et al. \cite{longchat2023} uncover that many open-source LLMs fall short of expectations in long-context scenarios, indicating a difference between having the architectural capacity for long inputs and actual performance efficacy.  Similarly, `Lost in the Middle: How Language Models Use Long Contexts` \cite{liu2023lost} authors find an unwanted U-shaped performance curve in LLMs, where the models process information well at the beginning or end of a context, but struggle with details in the middle. These results highlight a notable performance degradation as context windows expand and provide further motivation for work beyond merely extending the context window.

\paragraph{Remembering in long contexts.}
In order to mitigate these struggles, researchers have employed many strategies. Recent advancements in efficient attention mechanisms are addressing such limitations of processing long inputs. Techniques like Luna \cite{ma2021luna}, alternative attention mechanisms \cite{shen2020efficient}\cite{li2020linear}, Transient Global \cite{guo-etal-2022-longt5}, CoLT5 \cite{ainslie2023colt5}, Focused Transformer \cite{tworkowski2023focused}, and LM-INFINITE \cite{han2023lm} enhance LLMs' ability to handle extended sequences, with some claiming to enable up to 256k tokens. Innovations such as the Synthesizer \cite{pmlr-v139-tay21a}, fixed-size memory attention \cite{britz2017efficient}, and dilated attention \cite{ding2023longnet} further optimize processing efficiency and improve performance with significantly longer input contexts than base models.

Other researchers are exploring positional embedding schemes to enhance the ability of Transformer-based LLMs to generalize to longer sequence lengths than those encountered during training. Techniques like Absolute Positional Embeddings \cite{vaswani2023attention}, Relative Positional Embeddings \cite{shaw-etal-2018-self} \cite{huang2018music} \cite{dai-etal-2019-transformer} \cite{chen2023extending}, Rotary Position Embeddings \cite{su2023roformer}, and ALiBi \cite{press2022train} \cite{workshop2023bloom} aim to improve LLMs' handling of longer sequences by modifying how positional information is integrated. Despite these advancements, the effectiveness of these schemes in enabling LLMs to effectively remember and perform in long contexts varies, with some showing promising results in specific tasks, while others struggle with length generalization \cite{taylor2022galactica} \cite{kazemnejad2023impact} \cite{anil2022exploring}.

Beyond the scope of Transformers, researchers are exploring alternatives tailored for efficiently handling long contexts without relying on traditional attention and positional embeddings. Techniques like state space models (SSMs) \cite{fu2023hungry}, the Hyena operator \cite{poli2023hyena}, and the Block-State Transformer \cite{fathi2023blockstate} are being developed to process extended sequences with near-linear computational complexity, improving long-range recall and token comparison. Additionally, recurrent neural network (RNN) based approaches, such as the Receptance Weighted Key Value (RWKV) \cite{peng2023rwkv}, leverage RNN architectures, combining Transformer-like parallelization with RNN efficiency, to effectively manage long sequences in large-scale models.

\begin{figure*}[ht]
    \centering
    \includegraphics[width=\textwidth]{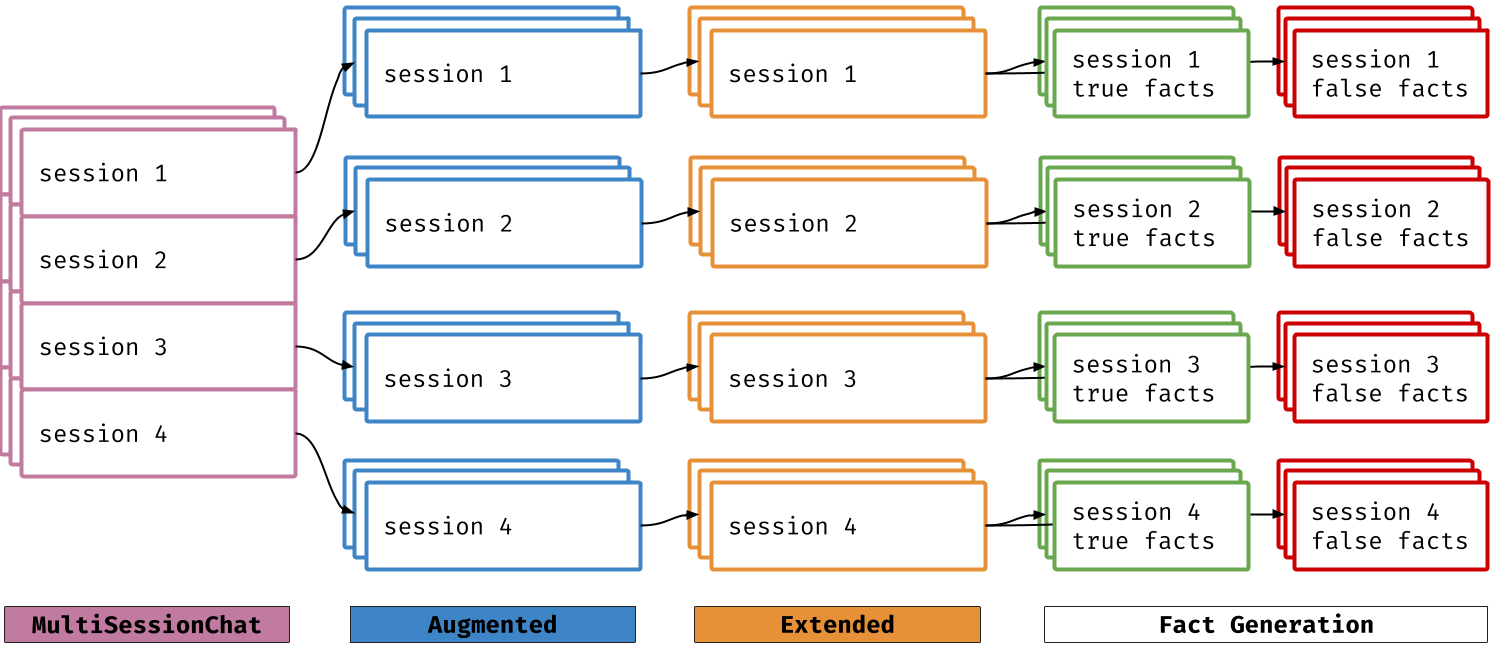}
    \caption{The creation pipeline from MSC to \name.  Each arrow represents a passage of text fed into GPT-4 along with the accompanying prompt from the \hyperref[Appendix A]{Appendix} for that step.}
    \label{fig:fig2}
\end{figure*}

\paragraph{Remembering beyond context window.}

The following papers are all proposals augmenting LLMs with some sort of memory solution.  Often, these systems will incorporate context that has occurred beyond the context window, crucial for the problem we are talking about.  

In many cases, the solution is some form of retrieval augmentation, as is discussed by Shuster \cite{shuster2021retrieval}. Zhu et al. \cite{zhu2023pose} introduced a novel method, Positional Skip-wise Training (PoSE), to efficiently extend LLMs' context window capabilities. 
 \cite{xu2023retrieval} discusses the idea of retrieval augmented generation and compares it directly to a longer context window, showing that similar performance can be achieved with both. \cite{jiang2023longllmlingua} focuses on an approach that performs prompt compression for longer contexts.  A few retrieval augmented generation techniques include a self-retrieval model \cite{rubin2023long}, a recursive summarization technique \cite{wang2023recursively}, a self controlled memory system \cite{liang2023unleashing} and the creation and querying of a memory tree \cite{chen2023walking}. \cite{wang2023augmenting} details their approach for memorization and use of long-term context.  Other work surrounds a long term memory mechanism \cite{zhong2023memorybank}, database read write memory mechanisms \cite{modarressi2023retllm} \cite{bertsch2023unlimiformer}, a knowledge graph for QA in conversational dialogue \cite{hixon2015learning}, a mixture of experts (MoE) augmented model \cite{santos2023memory} and an LLM specialized for web content \cite{gur2023real}.

It is clear that this problem is well recognized in the research community.  We applaud the work and efforts made by the various researchers to solve the memory problem of LLMs.  However, these proposed solutions all lack a benchmark evaluation methodology with which to prove their efficacy at remembrance in long contexts.

\paragraph{Existing dialog benchmarks.} Dialogue datasets with which to evaluate come in a variety of forms. The Persona-Chat dataset is one of the most well known datasets for dialogue \cite{zhang2018personalizing}. Turkers were given a set of personality traits and told to converse with another Turker based on their written personality. The result was a collection of short human-written dialogues. This work was extended into MultiSessionChat \cite{xu2021goldfish}, where two Turkers were given an original Persona-Chat entry and told to continue the conversation. The result is a human written, 4 session, up to 14 utterance dialogue dataset. While this is an increase in length, this is still too short for evaluating memory in long contexts. We find similar problems in all other dialogue datasets, where they do not include enough context to be constituted as ``long``\cite{ubuntudata} \cite{li2017dailydialog} \cite{sun2019dream} \cite{rashkin2019towards} \cite{dai-etal-2022-dialog} \cite{kim2022prosocialdialog} \cite{wu2022qaconv} \cite{lee2022dialogcc}.  Additionally, \cite{shuster2020the} presents dodecaDialogue which encompass 12 tasks to evaluate various abilities of a conversational agent to interact with humans.  All of these tasks are evaluated in short context settings and lack evaluation in long contexts.  The main goal of our work is to provide a long-context dialogue dataset and accompanying tasks to evaluate memory performance.  We present a comparison of \name{} to select related datasets in Table~\ref{tab:dialogue_datasets}.

\section{\name}

\name ~ is a long-context dialogue dataset, parsed into four distinct sessions between two users, with an accompanying list of true and false facts about each session.

\paragraph{MultiSessionChat.}
Building upon the PersonaChat \cite{zhang2018personalizing} framework, the MultiSessionChat \cite{xu2021goldfish} dialogue dataset represents an evolutionary step in dialogue data. Although it extends previous work, we recognize a need for even longer dialogues for our specific research. Notably, the dialogues in MultiSessionChat, despite their structured design, often appear somewhat artificial, lacking natural transitions between topics. Nevertheless, the dataset has certain attractive features. One key aspect is the division of conversations into four distinct sessions, which is crucial in pinpointing performance variations at different stages of the dialogue. Another significant factor is the human authorship of the dataset, which is vital for our evaluation of LLMs, as it provides a realistic benchmark for assessing their competency in mimicking human conversation.

\paragraph{Augmenting and extending.}
To achieve a long and harmonious dialogue, we utilize one of the current state of the art LLMs, GPT-4 \cite{openai2023gpt4}. The original MSC dialogue is augmented and extended in a multi-step pipeline before achieving the final result. We begin with improving fluency, in which we feed each ~250 token MSC session into the LLM with a carefully engineered prompt. This prompt requests GPT-4 to rewrite the original dialogue in order to add smoother transitions while retaining all of the key points. The prompt used for this can be found in \hyperref[Appendix A]{Appendix A}. The next step is to extend the length of the dialogue. Each of these augmented sessions are sent back into GPT-4 with a different prompt asking it to continue the conversation. This extends each session to $\sim 1000$ tokens. The prompt used for this can be found in \hyperref[Appendix B]{Appendix B}. The result is polished and extended dialogues, each $\sim 4,000$ token (4 sessions x 1,000 tokens) and originally written by humans. Thus, we achieve the desired human-written, long-context dialogues.

\paragraph{Fact bank.}
With the dialogues in place, the next step is to design a task to assess memory capabilities. We've selected fact retrieval as this task. Leveraging the extensive, natural dialogues, we instruct GPT-4 to craft both true and false statements about each character in every session. For authenticity, true statements are based strictly on explicitly mentioned facts, not on inferences. In parallel, we ask GPT-4 to deliberately create false statements, again relying only on explicit content. To validate these facts, we have the model cite the specific line number where each fact can be confirmed or refuted. This approach enables meticulous manual verification and enhances the model's ability to produce contextually accurate facts. These prompts are available in Appendices C and D.

\paragraph{} The result is \name, a long, realistic dialogue dataset, with true and false facts associated with each, and originally written by humans.  Source code for the entire process as well as generating more dialogues will be released in the future. Figure~\ref{fig:fig2} demonstrates the entire pipeline of dataset creation.

\paragraph{Statistics.}
Table~\ref{tab:stats} shows various statistics for \name. We use OpenAI's TikToken \cite{tiktoken2023} tokenizer to count tokens for the \texttt{gpt-3.5-turbo} model. In our framework, each fact presented in the dialogues is treated as an individual task, culminating in nearly 200 tasks per conversation to rigorously test memory capabilities. It's noteworthy that the conversations alone account for almost 4,000 tokens, and this figure significantly exceeds 4,000 when combined with key facts and the associated prompts.

\setlength{\heavyrulewidth}{1.5pt}
\begin{table}[ht]
\centering
\begin{tabular}{lc} 
\toprule
    {Statistic} & {Value} \\ 
\midrule
    \# of Conversations & 100 \\
    \# of Questions & 16K \\
    Avg \# tokens per conv & 3.7K \\
    Avg \# utterances per conv & 110.3\\
    Avg \# tokens per utterance & 34.1\\
    Avg \# facts per conv & 197.8\\
    Avg \# true facts per conv & 117.0\\
    Avg \# false facts per conv & 80.8\\
\midrule
    \# sessions per conv & 4\\
    Avg \# tokens per session & 936.4 \\
\bottomrule
\end{tabular}
\caption{Dataset Statistics for \name.}
\label{tab:stats}
\end{table}

\section{Experimental Setup: Benchmarking Memory with \name}

To rigorously benchmark memory in dialogue contexts, we have designed an experimental setup with \name, leveraging the dialogues and facts to create memory tasks.

\paragraph{Task Creation.}

In this setup, we randomly select 40 facts per session from the dataset. This randomization ensures a diverse and unbiased sample, crucial for a comprehensive assessment of memory capabilities. Each session's facts are then woven into ~\texttt{memory tasks}: starting with the dialogue, followed by a list of candidate facts specific to the session, and concluding with clear instructions for response.  We direct the model to identify the candidate facts it deems true based on the dialogue, and importantly, to also indicate any uncertainties.  This allowed for granularity in the responses with respect to the model self-reporting when it is unsure.  For a view of these instructions, refer to \hyperref[Appendix E]{Appendix E}. The approach slightly differs for human trials, as outlined in \hyperref[Appendix F]{Appendix F}.

\paragraph{Task Configuration.}

Following the task creation phase, we specifically configure these ~\texttt{memory tasks} to facilitate an in-depth evaluation of memory performance.  These tasks are designed with varying lengths, ranging from a single session (approximately 1,000 tokens) to eight sessions (approximately 8,000 tokens). This variability is for testing the memory capacity across different lengths of dialogue, both within and out of the context window.  For each configuration, we generate specific ~\texttt{memory tasks} relevant to the sessions included. For instance, in a setup involving four sessions, separate tasks are created for each session, yet all four are presented together in the dialogue context. This approach tests the model's memory at individual points and assesses its ability to retain and recall information across longer contexts.  For a 4,000 max token count LLM, the ~\texttt{memory tasks} with 1 through 4 sessions evaluate within context memory, while those with 5 through 8 sessions evaluate out of context memory.

This configuration enables a nuanced analysis of an entity's performance throughout the dialogue. By observing how well facts are remembered from different parts of the conversation – be it from the beginning, middle, or end – we gain insights into the strengths and limitations of LLMs in handling extended dialogue scenarios.

Furthermore, the unique nature of each dialogue in \name~ allows for even more extensive configurations. By concatenating multiple dialogues, we can create scenarios simulating extremely long conversations.  For example, a scenario where there are one hundred sessions present there would be approximately 100,000 tokens. This scalability provides a robust framework for testing memory in contexts that far exceed the standard capacities of current LLMs. Such extensive dialogues offer a stringent test bed, where questions about content from any session – like the thirteenth in a hundred – remain unique and verifiable.

When we are evaluating performance on a ~\texttt{memory task} that is too long to fit into a context window, we cut tokens from the beginning of the conversation until it fits. For example, if there is a 5,000 token dialogue to be processed by a 4,000 token context window, we remove the beginning 1,000 tokens.  As in base LLMs, once you surpass the token count, the context first seen is erased.  If we are evaluating with a model that has an augmented memory module but can only process certain amounts of tokens at a time, we input the conversation sequentially.  We choose to do this as it simulates exactly what happens to these chatbots beyond their context window.

\paragraph{Reasoning}

The crux of our experiment is the use of true/false fact identification as the primary method for memory evaluation. We favor this approach over traditional question-answering formats due to its definitive nature in assessing memory accuracy. True/false questions offer a clear metric to evaluate an LLM's ability to recall, understand context, and discern between accurate and inaccurate statements. This method provides a more direct measurement of memory, encompassing both recall and comprehension abilities, thus serving as an effective tool for our memory benchmarking objectives

The human-written dialogues of \name~ are pivotal to our study, enabling a direct comparison of LLM performance against human memory capabilities in a conversational context. The extensive length of these dialogues challenges LLMs to process and recall information over long contexts, mirroring real-life conversational memory demands. The dataset's structure, with distinct sessions, further allows for a granular analysis of memory performance across different stages of the dialogue.

The choice of 40 facts per session aligns with the average fact count generated per dialogue, providing a consistent and manageable task design. The token length of each task mirrors the typical 4,000-token capacity of popular LLMs, ensuring our tasks are representative of real-world LLM processing limits. The design inherently supports scalability, as multiple sessions and unique conversations can be concatenated, maintaining task integrity and relevance across a broader range of contexts.

Our selection of 100 conversations, resulting in approximately 370K tokens, is deliberate. This token count is longer than the longest context window, and can therefore test well beyond the maximum of memory capabilities in LLMs today.

\section{Quality Control and Human Trials}

To ensure the integrity and validity of the \name~ dataset and associated tasks, meticulous quality control measures were implemented at every stage of the data creation process. The precision in task formulation is crucial, as the reliability of our benchmarking for memory rests on the accuracy and clarity of the tasks presented. This involved iterative refinement of prompts, crafted to be highly specific and directive in nature. A thorough manual review of numerous dialogues was undertaken by members of our team to confirm the consistency and realism of the interactions, as well as the correctness of the accompanying facts.

For external validation of our dataset and tasks, human trials were conducted via Amazon Mechanical Turk. Participants were assigned ~\texttt{memory tasks} analogous to those designed for LLMs, with a focus on evaluating their ability to discern true from false statements based on the dialogue context. This dual-purpose approach not only verified the feasibility and accuracy of the tasks but also provided a comparative baseline between human memory capabilities and LLM performance.

To maintain high standards of response quality, which has been shown to be questionable in some cases \cite{Kennedy_Clifford_Burleigh_Waggoner_Jewell_Winter_2020}:

\begin{itemize}
    \item A control fact was embedded within each set of 40 facts, crafted to be evidently false and unrelated to the dialogue. This acted as a quality filter; incorrect responses to this control fact indicated a lack of diligent engagement with the task.
    \item Specific criteria were set for participant selection: We targeted 'Master' Workers with a HIT Approval Rate of over 95\%, based in the US, and with more than 5000 approved HITs. This ensured a pool of experienced and reliable participants.
    \item Participants were compensated at a generous rate well above the United States national minimum wage at \$3.50 per task, equating to an approximate hourly rate of \$15. This rate was based on the average time required to complete a task and aimed to attract diligent and thoughtful engagement.
\end{itemize}

The results from these human trials served a twofold purpose: validating the correctness and solvability of the tasks, and providing a human performance benchmark for memory tasks, highlighting the differences in memory capabilities between humans and LLMs.  This is a distinctive feature of \name; earlier work rarely validated data against human baselines.

\section{Results: Benchmarking Memory with \name}

We ran a multitude of experiments using \name.  To begin, we started with evaluating the memory of out-of-the-box models as our baseline and showing that there is much work to be done.  We then evaluated a proposed memory-augmented solution from the paper `Unleashing Infinite-Length Input Capacity for Large-scale Language Models with Self-Controlled Memory System` \cite{liang2023unleashing}, showing that these solutions do work but are far from perfected.  Finally, we conduct human trials which prove that the tasks can be accurately completed irrespective of length.

\paragraph{Baseline: Standard LLM Memory.}

\textit{Modern LLMs have good memory within context, are able to report when they do not know something, and lack performance out of context.} 

\begin{figure}[ht]
\centering
\includegraphics[width=\columnwidth]{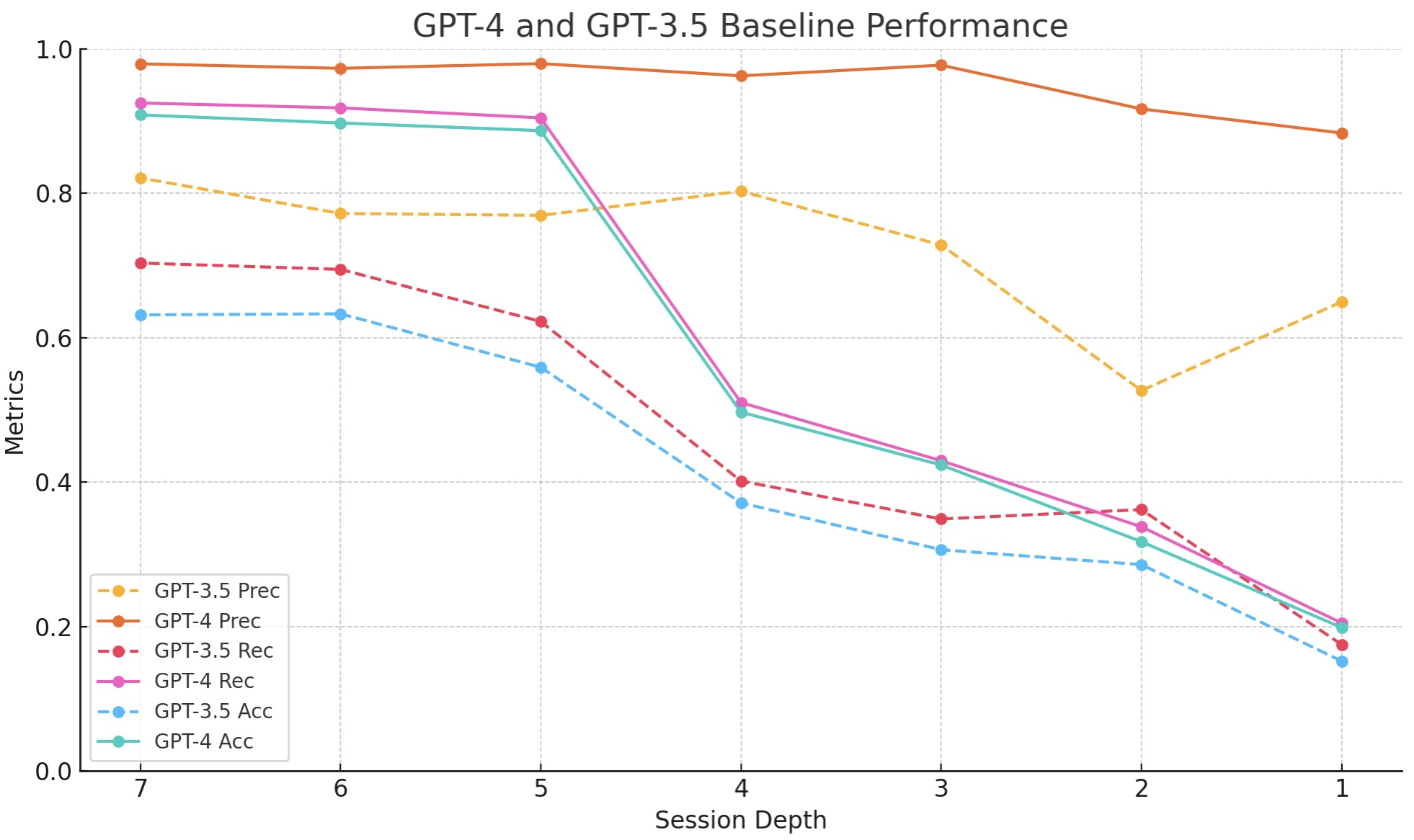}
\caption{GPT-3.5 and GPT-4 Baseline}
\caption*{\small Both models show a drastic drop in accuracy out-of-context. GPT-4's precision remains high even out-of-context, an indication of its lower hallucination rate.}
\label{fig:baseline-graph}
\end{figure}

In order to determine the memory capabilities of current standard LLMs, we conducted experiments on some popular open-sourced and private LLMs.  Both GPT-3.5 and GPT-4 had high marks across the board when evaluated on these memory tasks and they can see the context. However, as the dialogue gets bumped out of the context window, we see a sharp decline in performance.  This highlights the need for DiaLong, as it is a task that modern LLMs have yet to complete.  Interestingly, GPT-4 was able to identify quite well when it did not know the answer to a question, observed by high out-of-context precision.  This further emphasizes that the problem lies inherently in the model's memory rather than some form of reasoning and comprehension shortcoming.  As GPT-4 is considered one of the top performing state-of-the-art modern LLMs, this performance reflects all base out-of-the-box models without special memory augmentation to date.

\paragraph{Proposed Solutions: Enhanced LLM Memory.}

\textit{Memory augmented solutions improve LLM memory but have much more work to be done.}  
\begin{figure}[ht]
\centering
\includegraphics[width=\columnwidth]{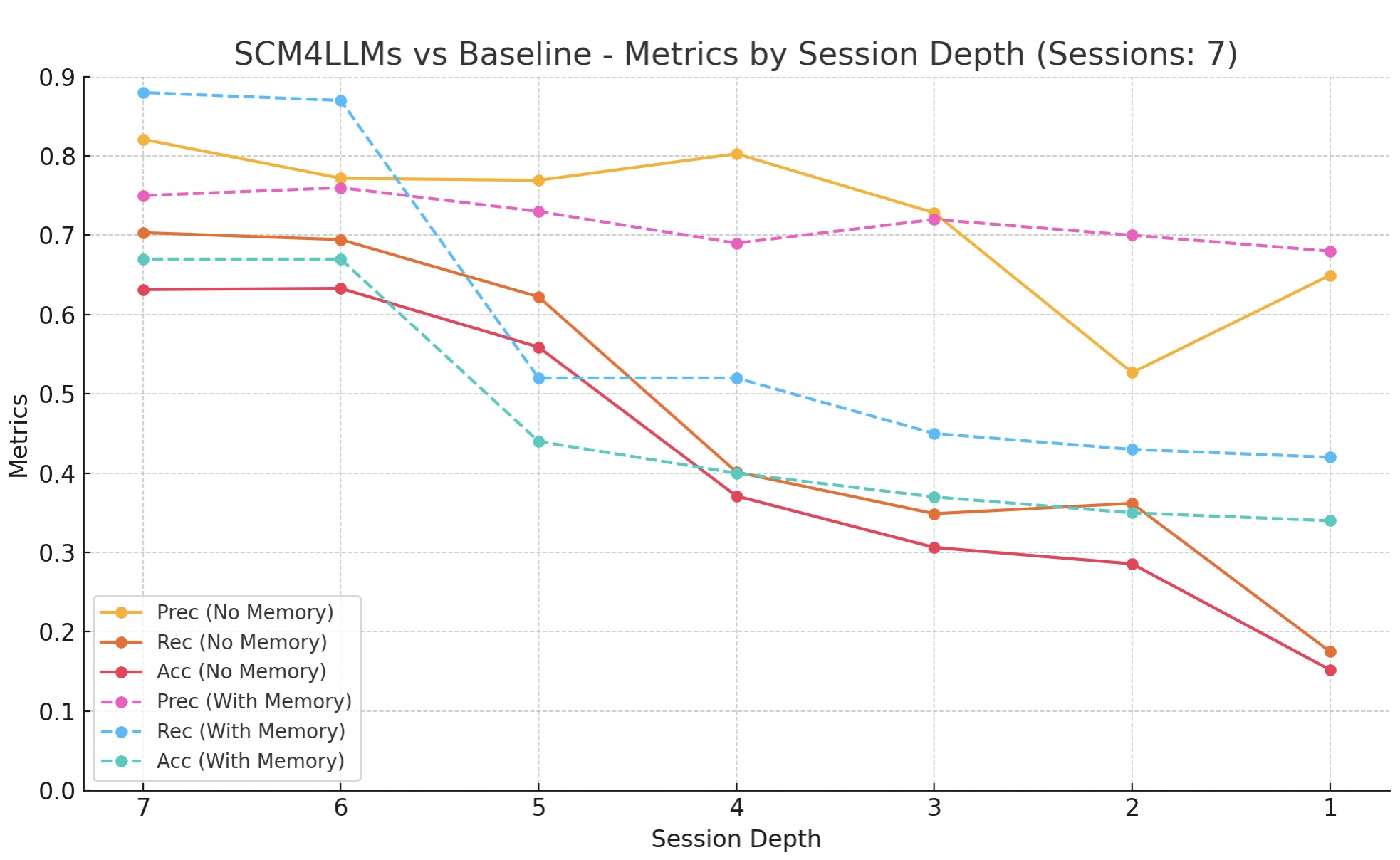}
\caption{GPT-3 Performance across session depth with and without the SCM4LLMs memory model.}
\captionsetup{justification=centering}
\caption*{\small Dashed lines = with memory. Solid lines = no memory. Metrics include precision, recall, and accuracy across 7 sessions. SCM4LLMs shows performance gains in out-of-context session depths.}
\label{fig:baseline-vs-memory}
\end{figure}

As discussed in the related works section, there are a multitude of proposed solutions to the memory problem of LLMs.  All of which, as noted, lack a rigorous evaluation on the memory ability.  We evaluated the memory-augmented solution SCM4LLMs \cite{liang2023unleashing} which powers GPT 3.5. The authors tested their model anecdotally, and did not have quantitative measures backing their solution.  This is at no fault to them, due to the fact that such a dataset did not exist before \name.  When evaluating the model on \name, we find that it exhibits significantly less of a drop in accuracy from within-context to out-of-context. Once the accuracy drops, it remains relatively consistent. This indicates that the memory model can retain some, but not all key facts from the conversation, and utilize them in the out-of-context scenario.

\paragraph{MTurk Comparison: Human Memory.}

\textit{Humans can perform just as well as LLMs at this task and are not limited by a context window.}  

The human trials demonstrated that this task is more than achievable by humans.  There is no drop in performance as the passages get longer, we just see that it takes the workers longer to complete.  We believe that this is due to the way that the humans are able to make use of the dialogue.  They have access to the entire conversation in order to reference and answer the questions.  This confirms that LLMs have a ways to go before achieving human-like memory performance on conversational dialogue tasks.

\begin{figure}[ht]
\centering
\includegraphics[width=\columnwidth]{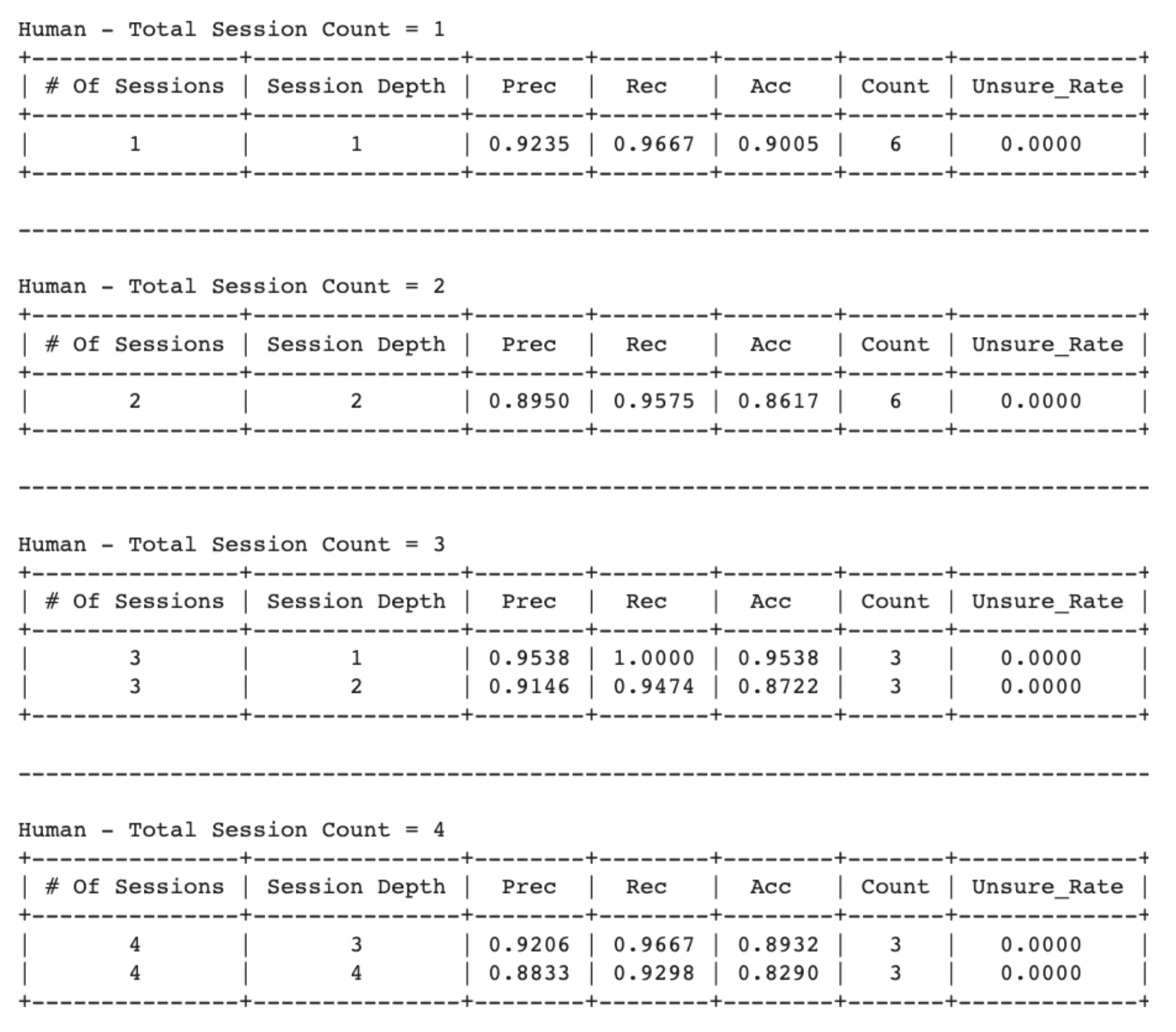}
\caption{Human Evaluation Performance}
\label{fig:human}
\end{figure}

\section{Conclusion}
We show the limitations of current LLMs in processing long dialogue contexts which are a significant barrier to achieving human-like conversational capabilities. We introduce \name, a long-context dialog dataset and evaluation framework, useful for benchmarking the memory performance of LLMs in extended dialogues. Our findings reveal a need for further advancements in the memory domain, emphasizing the importance of models that can sustain coherent, accurate, and engaging conversations over long contexts.

\bibliography{ref}
\bibliographystyle{acl_natbib}

\clearpage
\onecolumn
\appendix
{\LARGE\textbf{Supplemental Material}}
\section{Augmentation}
\vspace{-0.5cm} %
\label{Appendix A}
\begin{tcolorbox}[colback=red!5!white,colframe=red!75!black,title=\textbf{Augmentation Prompt:}]
\tt

Rework an existing dialogue between two characters, making it sound more natural, fluid, and human-like (such that it is indistinguishable from a human conversation).  The conversation is split into four distinct sections, each representing a different passage of time (ex. 2 hours later), but the characters and their personalities remain the same. Each session is separated by new lines. Remember, the conversation should be indistinguishable from that of a human.  While reworking the conversation, please ensure the character personas remain consistent throughout and change the character names to Person 1 and Person 2 (if they aren't already). Indicate where each session begins as the input indicates it.  Here is the dialogue to be reworked (between triple quotes): ``` \{conversation\} ```  

\end{tcolorbox}
\section{Extension}
\vspace{-0.5cm} %
\label{Appendix B}
\begin{tcolorbox}[colback=red!5!white,colframe=red!75!black,title=\textbf{Extension Prompt:}]
\tt
I have a conversation between two people that is split into 4 sessions.  I will give you the conversation for one of the sessions. Lengthen this session's conversation as much as you possibly can. While doing so, ensure that: - You continue the conversation from where it left off (do not repeat the given part of the session) - The conversation remains realistic — it is two human characters having a natural, fluid conversation - The character personas remain consistent throughout, even though the topics they discuss change - The dialogue extension is between 400 and 550 words - Never stop the dialogue midway. It should end at a reasonable point without me having to ask you to continue generating. Formatting: Start with triple quotes (```), then the dialogue extension, and end with triple quotes (```). No other output should be generated (ex. no ``Session extension: ``, ``extension: ``, etc.) Here is the conversation for the session (between triple quotes): ``` \{conversation\} ```
\end{tcolorbox}

\section{True Facts}
\label{Appendix C}
\begin{tcolorbox}[ colback=red!5!white,colframe=red!75!black,title=\textbf{True Facts Prompt:}]
\tt
Given the conversation below, list out all the facts that we can collect about Person {person\_num}. I want facts that have been explicitly said, not inferred. These facts can range from what the person thinks about a certain topic, the person's name, family relationships, interests, educational/career interests, travel experiences, their culture and so on. Do not provide any other explanation but just give the list in bullet points. Explicitly state that the fact is about Person {person\_num} when listing it out.  Do not write anything before the output of facts such as ``Sure! Here are the facts about Person {person\_num} that we can gather from the conversation:`` or ``Facts about Person {person\_num}:`` Just get to the point. The output should be a simple list of facts separated with a - and at the end of each fact, it should be separated with a new line. If there is no information on a given topic I mentioned, you do not need to output anything about it. Also, you do not need to ask me if there is any clarification. Only output the list as well as the line number you found the information from in parentheses. An example of an element of a list would look like this. - Person {person\_num} thinks that retro foods like peanut butter and banana toast or celery with peanut butter and raisins could be popular. (2) Dialogue (between triple quotes): ```\{conversation\}```
\end{tcolorbox}

\section{False Facts}
\label{Appendix D}
\begin{tcolorbox}[colback=red!5!white,colframe=red!75!black,title=\textbf{False Facts Prompt:}]
\tt
Below is a conversation between Person 1 and Person 2 I want you to come up with a list of false facts about Person {person\_num} that are on topic with what the conversation is about but wrong factually. Come up with at least 10 false facts about Person {person\_num}, for each false fact, have line numbers in () showing where the topic is mentioned.  I want facts that have been explicitly said, not inferred and then falsified. Do not provide any other explanation but just give the list in bullet points. Explicitly state that the false fact is about Person {person\_num} when listing it out, but do not state that it is false. Do not write anything before the output of facts such as ``Sure! Here are the false facts about Person {person\_num} that we can gather from the conversation:`` or ``False facts about Person {person\_num}:`` Just get to the point. The output should be a simple list of facts separated with a - and at the end of each fact, it should be separated with a new line. Also, you do not need to ask me if there is any clarification.  Only output the list as well as the line number you found the information from in parentheses. An example of an element of a list would look like this. - Person {person\_num} thinks that retro foods like peanut butter and banana toast or celery with peanut butter and raisins could be popular. (2) Dialogue (between triple quotes): ```\{conversation\}```
\end{tcolorbox}

\section{Memory Task - LLM}
\label{Appendix E}
\begin{tcolorbox}[colback=red!5!white,colframe=red!75!black,title=\textbf{Memory Task Prompt - LLM:}]
\tt
Here is a conversation between two people as well as a list of candidate true and false facts from the passage. Here is the conversation: ```{conversation}``` And here are the candidate facts that may be true or false \{facts\_string\}. Given what you understand from the conversation give me a list of the candidate fact numbers you think are true. Strictly respond with a list of numbers in brackets in brackets [] associated to the line numbers of true candidate facts separated by commas. Do not say any other words or explanation. For example if you think facts 91, 92, and 93 are true please respond [91, 92, 93]. If you cannot determine please return an empty list [].
\end{tcolorbox}

\section{Memory Task Prompt - Human}
\label{Appendix F}
\begin{tcolorbox}[colback=red!5!white,colframe=red!75!black,title=\textbf{Memory Task Prompt - Human:}]
\tt
Here is a conversation between two people as well as a list of candidate true and false facts from the passage. Here is the conversation: ```{conversation}``` And here are the candidate facts that may be true or false \{facts\_string\}. Given what you understand from the conversation, please indicate whether you think each of the candidate facts are true or false. You may reference the passage as much as needed to select your answers.
\end{tcolorbox}

\section{Amazon Mechanical Turk Task}
\label{Appendix G}

For the Mtruk experiment, we utilized a prompt that was nearly identical to the model. The only difference was that we included additional instructions and examples to help participants better understand the purpose of the experiment. Additionally, we added a deliberately incorrect and irrelevant option to ensure that participants were carefully reading all the content.

\begin{figure}[ht]
\centering
\includegraphics[width=\columnwidth]{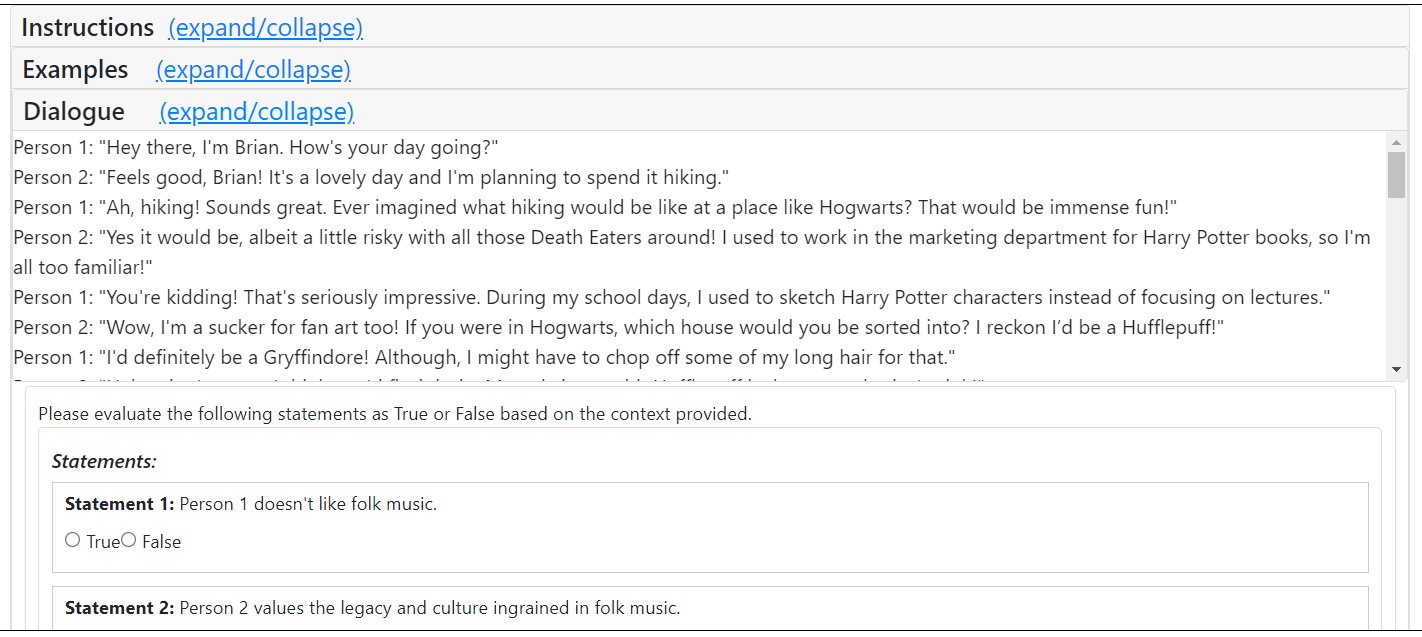}
\caption{The basic layout for Mturk experiment}
\label{fig:mturk}
\end{figure}

\pagebreak
\section{Open Source Model Details}
\label{Appendix H}

For open source models, we evaluated them with the same basic prompt. For each model, we only did slight modifications on the prompt. Also, 
In order to maintain the consistency and accuracy, we set temperatrue=0.3 and top\_p = 0.95.

\begin{tcolorbox}[colback=gray!5!white,colframe=gray!75!black,title=\textbf{Basic Prompt:}]
\tt

Here is a conversation between two people as well as a list of candidate true and false facts from the passage. Here is the conversation:

\textbf{\{Dialogue Context\}}

And here are the candidate facts that may be true or false:

\textbf{\{Fact Banks\}}

Given what you understand from the conversation give me a list of the candidate fact numbers you think are true. Strictly respond with a list of numbers in brackets in brackets [] associated to the line numbers of true candidate facts separated by commas.  Do not say any other words or explanation.  For example if you think facts 91, 92, and 93 are true please respond [91, 92, 93].  If you cannot determine please return an empty list [].  

\end{tcolorbox}

Here is the prompt for LLaMa2:

\begin{tcolorbox}[colback=gray!5!white,colframe=gray!75!black,title=\textbf{LLaMa2 Prompt:}]
\tt

<<SYS>>

You're are a helpful Assistant, and you only response to the "Assistant"
Remember, maintain a natural tone. Be precise, concise, and casual. Keep it short

<</SYS>>

[INST]

User:\textbf{\{Basic Prompt\}}

[/INST]

Assistant:

\end{tcolorbox}

Here is the prompt for Xwin:

\begin{tcolorbox}[colback=gray!5!white,colframe=gray!75!black,title=\textbf{Xwin Prompt:}]
\tt

A chat between a curious user and an artificial intelligence assistant. The assistant gives helpful, detailed, and polite answers to the user's questions. 

USER: \textbf{\{Basic Prompt\}}

ASSISTANT: 

\end{tcolorbox}

Here is the prompt for Tulu2:

\begin{tcolorbox}[colback=gray!5!white,colframe=gray!75!black,title=\textbf{Tulu2 Prompt:}]
\tt

<|user|>

\textbf{\{Basic Prompt\}}

<|assistant|>

\end{tcolorbox}

\pagebreak
\section{Private Model Details}
\label{Appendix I}

We use the same basic prompt as open source models. The only differences are still some minor modifications on Claude V2. Also, the temperature=0.3 and top\_p=0.95.

\begin{tcolorbox}[colback=gray!5!white,colframe=gray!75!black,title=\textbf{Claude V2 Prompt:}]
\tt

Human:

\textbf{\{Basic Prompt\}}

Assistant:

\end{tcolorbox}

\section{LLMLingua Details}
\label{Appendix J}

For LLMLingua, we compress only the Dialogue\_Context and adjust the target\_token to a specific numerical value, ensuring that the entire prompt remains under 4,000 tokens. This approach guarantees that it fits within the maximum token limit allowed by any open-source model.

\end{document}